%% file: main_arxiv.tex
\documentclass[10pt,twocolumn,letterpaper]{article}

\usepackage{iccv}
\usepackage{times}
\usepackage{epsfig}
\usepackage{graphicx}
\usepackage{amsmath}
\usepackage{amssymb}
\usepackage{amsthm}

\usepackage{booktabs}
\usepackage{multirow}

\usepackage{mathtools}
\DeclarePairedDelimiter{\ceil}{\lceil}{\rceil}

\usepackage{kotex}

\usepackage[pagebackref=true,breaklinks=true,letterpaper=true,colorlinks,bookmarks=false]{hyperref}

\iccvfinalcopy

\ificcvfinal\fi

\usepackage{xcolor}
\usepackage{xspace}

\newcommand{\agis}{AGIS-Net~\cite{gao2019agisnet}\xspace}
\newcommand{\funit}{FUNIT~\cite{liu2019funit}\xspace}

\newcommand{\lffont}{LF-Font~\cite{park2021lffont}\xspace}

\newcommand{\ours}{MX-Font\xspace}
\newcommand{\methodname}{Multiple  Localized Experts Few-shot Font Generation Network (\ours)\xspace}
\newcommand{\methodnameinit}{{\bf M}ultiple  Localized e{\bf X}perts Few-shot {\bf Font} Generation Network (\ours)\xspace}
\definecolor{darkergreen}{RGB}{21, 152, 56}
\definecolor{red2}{RGB}{252, 54, 65}

\newcommand{\supp}{supplementary materials\xspace}

\newcommand{\ffg}{FFG\xspace}

\usepackage{pifont}
\newcommand{\yesmark}{\textcolor{darkergreen}{\ding{52}}}
\newcommand{\nomark}{\textcolor{red2}{\ding{56}}}

\begin{document}

\title{Multiple Heads are Better than One:\\ Few-shot Font Generation with Multiple Localized Experts}

\author{Song Park\textsuperscript{\rm 1} \quad Sanghyuk Chun\textsuperscript{\rm 2, 3} \quad Junbum Cha\textsuperscript{\rm 3} \quad Bado Lee\textsuperscript{\rm 3} \quad Hyunjung Shim\textsuperscript{\rm 1} \\
\\
\textsuperscript{\rm 1} School of Integrated Technology, Yonsei University \quad
\textsuperscript{\rm 2} NAVER AI Lab \quad
\textsuperscript{\rm 3} NAVER CLOVA
}

\maketitle
\ificcvfinal\thispagestyle{empty}\fi

\begin{abstract}
A few-shot font generation (FFG) method has to satisfy two objectives: the generated images should preserve the underlying global structure of the target character and present the diverse local reference style. Existing FFG methods aim to disentangle content and style either by extracting a universal representation style or extracting multiple component-wise style representations. However, previous methods either fail to capture diverse local styles or cannot be generalized to a character with unseen components, e.g., unseen language systems. To mitigate the issues, we propose a novel FFG method, named Multiple Localized Experts Few-shot Font Generation Network (MX-Font). MX-Font extracts multiple style features not explicitly conditioned on component labels, but automatically by multiple experts to represent different local concepts, e.g., left-side sub-glyph. Owing to the multiple experts, MX-Font can capture diverse local concepts and show the generalizability to unseen languages. During training, we utilize component labels as weak supervision to guide each expert to be specialized for different local concepts. We formulate the component assign problem to each expert as the graph matching problem, and solve it by the Hungarian algorithm. We also employ the independence loss and the content-style adversarial loss to impose the content-style disentanglement. In our experiments, MX-Font outperforms previous state-of-the-art FFG methods in the Chinese generation and cross-lingual, e.g., Chinese to Korean, generation. Source code is available at \url{https://github.com/clovaai/mxfont}.
\end{abstract}

\section{Introduction}

A few-shot font generation task (\ffg)~\cite{sun2018_ijcai_savae, zhang2018_cvpr_emd, gao2019agisnet, srivatsan2019_emnlp_deepfactorization, cha2020dmfont, cha2020dmfontw, park2021lffont} aims to generate a new font library using only a few reference glyphs, \eg, less than 10 glyph images, without additional model fine-tuning at the test time. \ffg is especially a desirable task when designing a new font library for glyph-rich scripts, \eg, Chinese ($>$ 50K glyphs), Korean ($\approx$ 11K glyphs), or Thai ($\approx$ 11K glyphs). It is because the traditional font design process is very labor-intensive due to the complex characteristics of the font domain. Another real-world scenario of \ffg is to extend an existing font design to different language systems. For example, an international multi-media content, such as a video game or movie designed with a creative font, is required to re-design coherent style fonts for different languages.

\begin{figure}
    \centering
    \includegraphics[width=\linewidth]{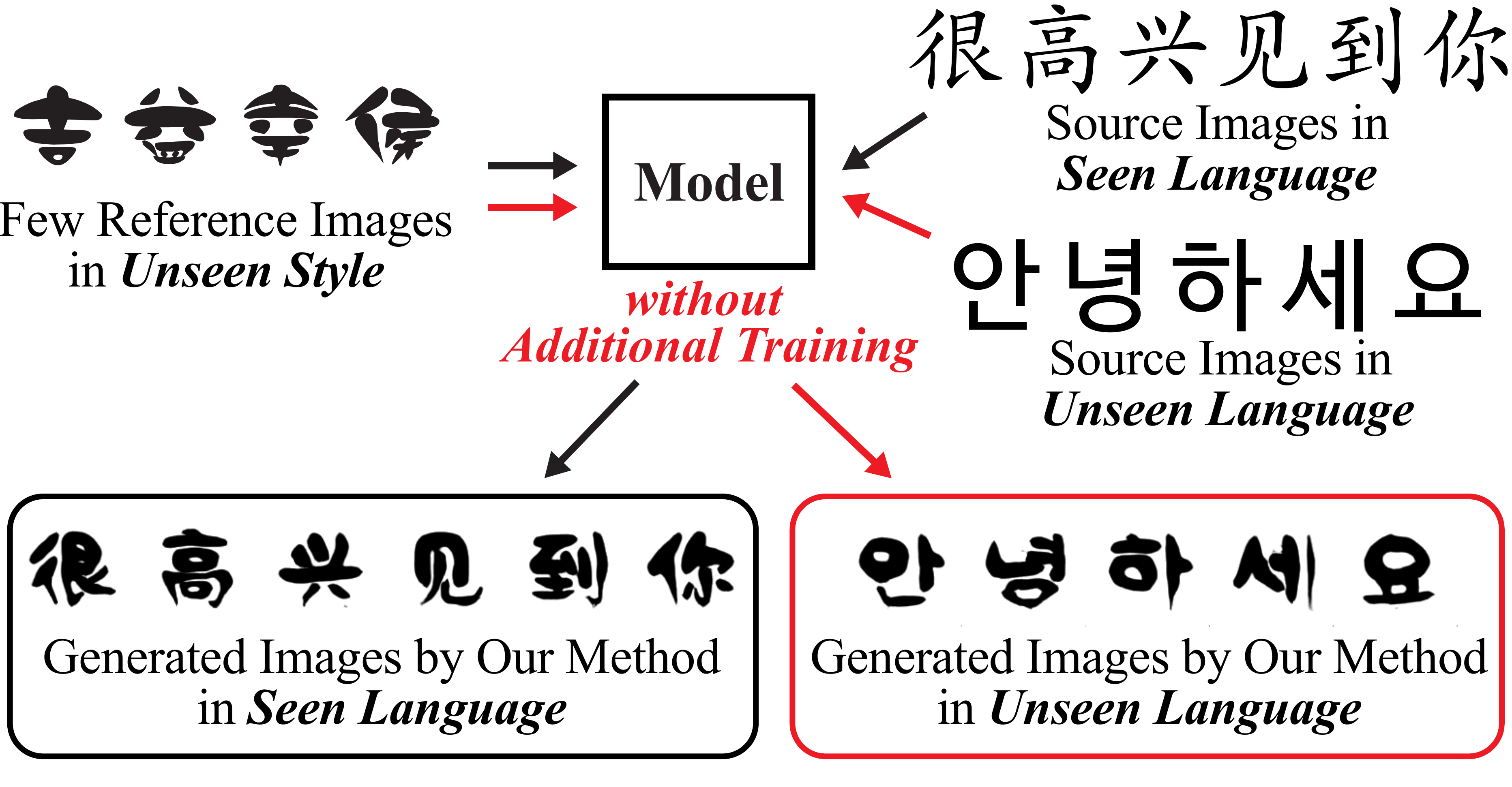}
    \caption{\small {\bf Cross-lingual few-shot font generation results by \ours.} With only four references, the proposed method, \ours, can generate a high quality font library. Furthermore, we first show the effectiveness of the proposed method on the {\it zero-shot cross-lingual} few-shot generation task, \ie, generating unseen Korean glyphs using the Chinese font generation model.}
    \label{fig:teaser}
    \vspace{-1em}
\end{figure}

\begin{figure*}
    \centering
    \includegraphics[width=\linewidth]{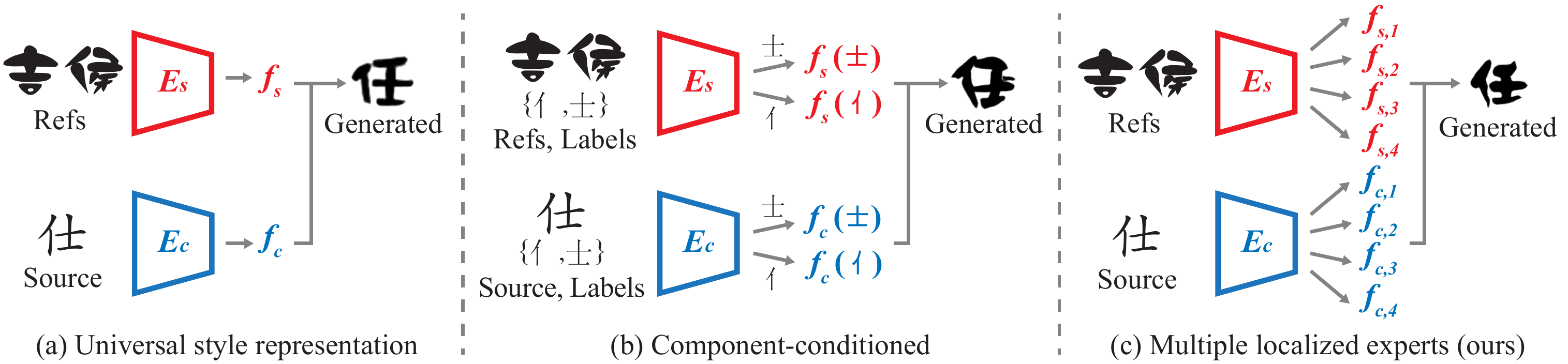}
    \caption{\small {\bf Comparison of \ffg methods.} Three different groups of \ffg are shown. All methods combine style representation $f_s$ from a few reference glyphs (Refs) by a style encoder ($E_s$) and content representation $f_c$ from a source glyph (Source) by a content encoder ($E_c$). (a) Universal style representation methods extract only a single style feature for each font. (b) Component-conditioned methods extract component conditioned style features to capture diverse local styles (c) Multiple localized experts method (ours) generates multiple local features without an explicit condition, but attends different local information of the complex input glyph. The generated images in (a), (b) and (c) are synthesized by \agis, \lffont and \ours, respectively.}
    \label{fig:concept_comparison}
    \vspace{-.5em}
\end{figure*}

A high-quality font design is obliged to satisfy two objectives. First, the generated glyph should maintain all the detailed structure of the target character, particularly important for glyph-rich scripts with highly complex structure. For example, even very small damages on a local component of a Chinese glyph can hurt the meaning of the target character. As another objective, a generated glyph should have a diverse local style of the reference glyphs, \eg, serif-ness, strokes, thickness, or size. To achieve both objectives, existing methods formulate \ffg by disentangling the content information and the style information from the given glyphs~\cite{sun2018_ijcai_savae, zhang2018_cvpr_emd, gao2019agisnet, cha2020dmfont, park2021lffont}. They combine the content features from the source glyph and the style features from the reference glyphs to generate a glyph with the reference style. Due to the complex nature of the font domain, the major challenge of \ffg is to correctly disentangle the global content structure and the diverse local styles. However, as shown in our experiments, we observe that existing methods are insufficient to capture diverse local styles or to preserve the global structures of unseen language systems.

We categorize existing \ffg methods into {\it universal style representation methods}~\cite{sun2018_ijcai_savae, zhang2018_cvpr_emd, liu2019funit, gao2019agisnet} and {\it component-conditioned methods}~\cite{cha2020dmfont, park2021lffont}. Universal style representation methods~\cite{sun2018_ijcai_savae, zhang2018_cvpr_emd, liu2019funit, gao2019agisnet} extract only a single style representation for each style -- see Figure~\ref{fig:concept_comparison} (a). As glyph images are highly complex, these methods often fail to capture diverse local styles. To address the issue, component-conditioned methods~\cite{cha2020dmfont, park2021lffont} utilize {\it compositionality}; a character can be decomposed into a number of sub-characters, or {\it components} -- see Figure~\ref{fig:concept_comparison} (b). They explicitly extract component-conditioned features, beneficial to preserve the local component information. Despite their promising performances, their encoder is tightly coupled with specific component labels of the target language domain, which hinders processing the glyphs with unseen components or conducting a cross-lingual font generation.

In this paper, we propose a novel few-shot font generation method, named \methodnameinit, which can capture multiple local styles, but not limited to a specific language system. \ours has a multi-headed encoder, named {\it multiple localized experts}. Each localized expert is specialized for different local sub-concepts from the given complex glyph image. Unlike component-conditioned methods, our experts are not explicitly mapped to a specific component, but each expert implicitly learns different local concepts by weak supervision \ie component and style classifiers. To prevent that different experts learn the same local component, we formulate the component label allocation problem as a graph matching problem, optimally solved by the Hungarian algorithm~\cite{kuhn1955hungarian} (Figure~\ref{fig:localized_expert_concept}). We also employ the {\it independence loss} and the {\it content-style adversarial loss} to enforce the content-style disentanglement by each localized expert. Interestingly, with only weak component-wise supervision (\ie image-level not pixel-level labels), we observe that each localized expert is specialized for different local areas, \eg, attending the left-side of the image (Figure~\ref{fig:expert_cam}). While we inherit the advantage of component-conditioned methods~\cite{cha2020dmfont, park2021lffont} by introducing the multiple local features, our method is not limited to a specific language by removing the explicit component dependency in extracting features. Consequently, \ours outperforms the state-of-the-art \ffg in two scenarios: {\it In-domain transfer scenario}, training on Chinese fonts and generating an unseen Chinese font, and {\it zero-shot cross-lingual transfer scenario}, training on Chinese fonts and generating a Korean font. Our ablation and model analysis support that the proposed modules and optimization objectives are important to capture multiple diverse local concepts.

\begin{figure*}        
    \centering
    \includegraphics[width=\linewidth]{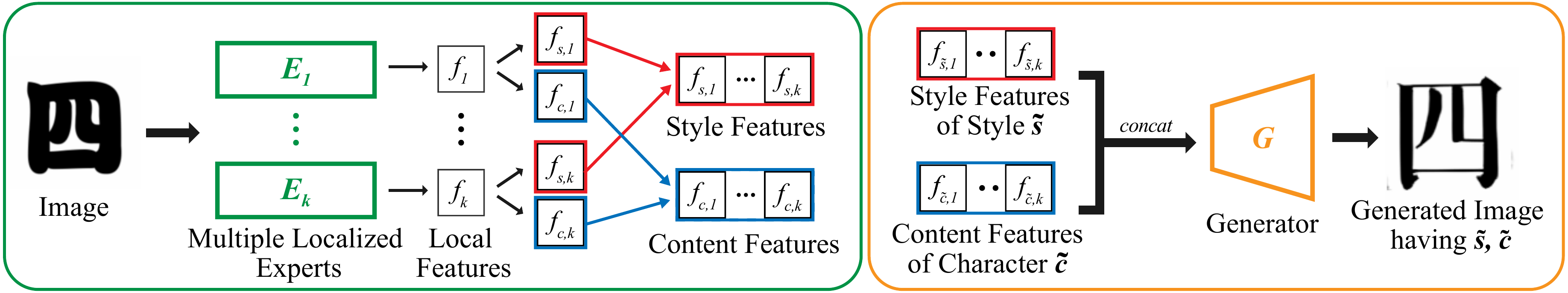}
    \caption{\small {\bf Overview of \ours.} Two modules of \ours used for the generation are described. The {\it multiple localized experts} (green box) consist of $k$ experts. $E_i$ (\ie $i$-th expert) encodes the input image to the $i$-th local feature $f_i$ and the $i$-th style and content feature $f_{s,i}$, $f_{c,i}$ are computed from $f_i$. The right yellow box shows how the generator $G$ generates the target image. When $k$ style features representing the target style $\widetilde{s}$ and $k$ content features representing the target style $\widetilde{c}$ are given, the target glyph having style $\widetilde{s}$ and character $\widetilde{c}$ is generated by passing the element-wisely concatenated style and content features to the $G$.}
    \label{fig:method_overview}
\end{figure*}

\section{Related Works}

\noindent\textbf{Style transfer and image-to-image translation.}
Few-shot font generation can be viewed as a task that transfers reference font style to target glyph. However, style transfer methods~\cite{gatys2016neuralstyle, adain, wct, deepphotostyle, photowct, wct2} regard the texture or color as a style while in font generation scenario, a style is often defined by a local shape, \eg, stroke, size, or serif-ness. On the other hand, image-to-image translation (I2I) methods ~\cite{isola2017_cvpr_pix2pix, zhu2017_iccv_cyclegan,stargan, liu2018unified, yu2019multi, starganv2} learn the mapping between domains from the data instead of defining the style. For example, FUNIT~\cite{liu2019funit} aims to translate an image to the given reference style while preserving the content. Many \ffg methods, thus, are based on I2I framework.

\noindent\textbf{Many-shot font generation methods.}
Early font generation methods, such as zi2zi~\cite{zi2zi}, aim to train the mapping between different font styles. A number of font generation methods~\cite{jiang2019_aaai_scfont,gao2020_aaai_chirogan,huang2020_eccv_rdgan,wu2020calligan} first learn the mapping function, and fine-tune the mapping function for many reference glyphs, \eg 775~\cite{jiang2019_aaai_scfont}. Despite their remarkable performances, their scenario is not practical because collecting hundreds of glyphs with a coherent style is too expensive. In this paper, we aim to generate an unseen font library without any expensive fine-tuning and collecting a large number of reference glyphs for a new style.

\noindent\textbf{Few-shot font generation methods.}
Since font styles are highly complex and fine-grained, utilizing statistical textures as style transfer is challenging. Instead, the majority of \ffg methods aims to disentangle font-specific style and content information from the given glyphs~\cite{zhang2018_cvpr_emd,srivatsan2019_emnlp_deepfactorization,azadi2018mcgan,gao2019agisnet,sun2018_ijcai_savae,li2021_wacv_ftransgan}. We categorize existing \ffg methods into two different categories. The universal style representation methods, such as EMD~\cite{zhang2018_cvpr_emd}, AGIS-Net~\cite{gao2019agisnet}, synthesize a glyph by combining the style vector extracted from the reference set, and the content vector extracted from the source glyph. \ours employs multiple styles, and does not rely on the font specific loss design, \eg, the local texture refinement loss by AGIS-Net~\cite{gao2019agisnet}. However, the universal style representation shows limited performances in capturing localized styles and content structures. To address the issue, {\it component-conditioned methods} such as DM-Font~\cite{cha2020dmfont}, LF-Font~\cite{park2021lffont}, remarkably improve the stylization performance by employing localized style representation, where the font style is described multiple localized styles instead of a single universal style. However, these methods require explicit component labels (observed during training) for the target character even at the test time. This property limits practical usages such as cross-lingual font generation. Our method inherits the advantages from component-guided multiple style representations, but does not require the explicit labels at the test time.

\section{Method}

We introduce a novel few-shot font generation method, namely \methodname. \ours has a multi-headed encoder called {\it multiple localized experts}, where $i$-th head (or expert $E_i$) encodes a glyph image $x$ into a local feature $f_i = E_i(x)$ (\S\ref{sec:arch}). We induce each expert $E_i$ to attend different local concepts, guided by a set of component labels $U_c$ for the given character $c$ (\S\ref{sec:best_allocation}). From $f_i$, we compute a local content feature $f_{c,i}$ and a local style feature $f_{s,i}$ (\S\ref{sec:cs_disentangle}). Once \ours is trained, we generate a glyph $\widetilde x$ with a character label $\widetilde c$ and a style label $\widetilde s$ by combining expert-wise features $f_{\widetilde c, i}$ and $f_{\widetilde s, i}$, from the source glyph and the reference glyph, respectively. (\S\ref{sec:generation}).

\subsection{Model architecture}
\label{sec:arch}
\begin{figure}
    \centering
    \includegraphics[width=\linewidth]{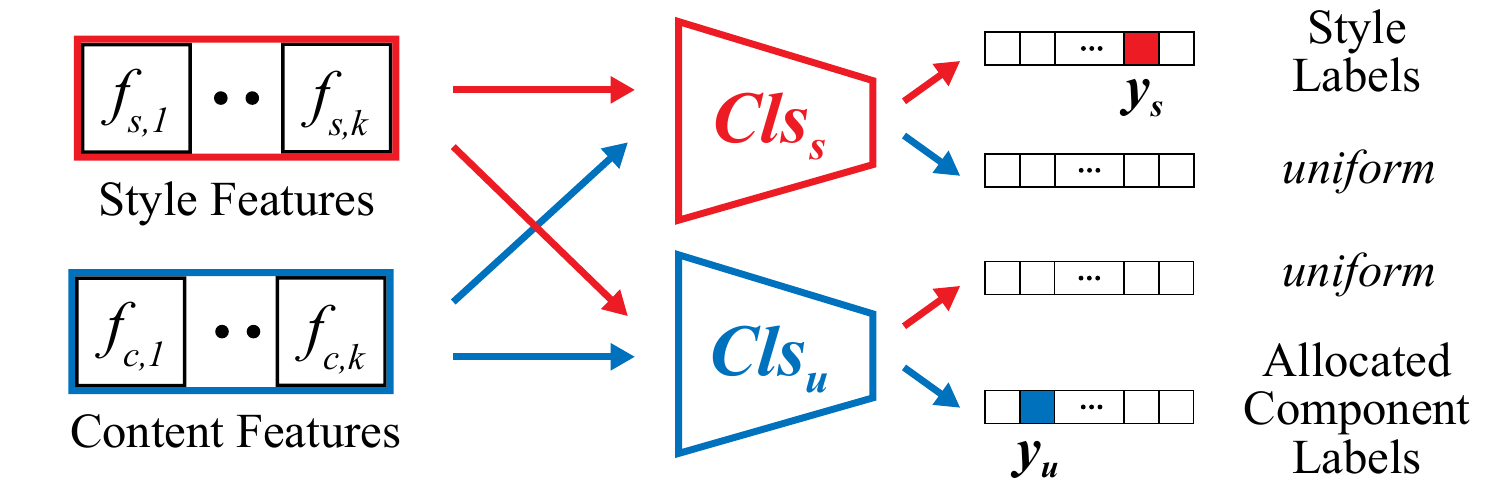}
    \caption{\small {\bf Feature classifiers.} Two feature classifiers, $Cls_s$ and $Cls_u$ are used during the training. $Cls_s$ classifies the style features to their style label $y_s$ while $Cls_u$ predicts the uniform probability from them. Similarly, $Cls_u$ classifies the content features to their allocated component labels $y_u$ while $Cls_s$ is fooled by them. The details are described in \S~\ref{sec:best_allocation} and \S~\ref{sec:cs_disentangle}.}
    \label{fig:feature_cls}
    \vspace{-0.5em}
\end{figure}

Our method consists of three modules; 1) $k$-headed encoder, or localized experts $E_i$, 2) a generator $G$, and 3) style and component feature classifiers $Cls_s$ and $Cls_u$. We illustrate the overview of our method in Figure~\ref{fig:method_overview} and Figure~\ref{fig:feature_cls}. We provide the details of the building blocks in the \supp.

The green box in Figure~\ref{fig:method_overview} shows how the {\it multiple localized experts} works. The {\bf localized expert} $E_i$ encodes a glyph image $x$ into a local feature $f_i = E_i(x) \in \mathbb R^{d \times w \times h}$, where $d$ is a feature dimension, and $\{w, h\}$ are spatial dimensions. By multiplying two linear weights $W_{i, c}, W_{i, s} \in \mathbb R^{d \times d}$ to $f_i$, a local content feature $f_{c,i} = W_{i, c}^\top f_i$ and a local style feature $f_{s,i} = W_{i, s}^\top f_i$ are computed. Here, our localized experts are not supervised by component labels to obtain $k$ local features $f_1, \ldots, f_k$; our local features are not component-specific features. We set the number of the localized experts, $k$, to 6 in our experiments if not specified.

\begin{figure}
    \centering
    \includegraphics[width=.9\linewidth]{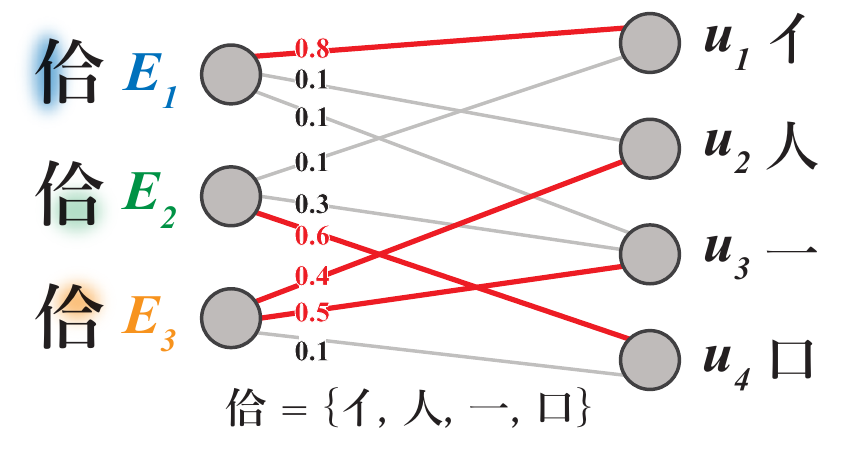}
    \caption{\small {\bf An example of localized experts.} The number of experts $k$ is three ($E_1, E_2, E_3$), and the number of target component labels $m$ is four ($u_1, \ldots, u_4$). An edge between an expert $E_i$ and a component $u_j$ means the prediction probability of $u_j$ by $E_i$ using the component classifier $Cls_u$. Our goal is to find a set of edges that maximizes the sum of predictions, where the number of the selected edges are upper bounded by $\max(k, m) = 4$ in this example. The red edges illustrate the optimal solution.}
    \label{fig:localized_expert_concept}
    \vspace{-.5em}
\end{figure}

We employ two {\bf feature classifiers}, $Cls_s$ and $Cls_u$ to supervise $f_{s, i}$ and $f_{c, i}$, which serve as weak supervision for $f_i$. The classifiers are trained to predict the style (or component) labels, thereby $E_i$ receives the feedback from the $Cls_s$ and $Cls_u$ that $f_{s, i}$ and $f_{c, i}$ should preserve label information. These classifiers are only used during training but independent to the model inference itself. Following the previous methods~\cite{cha2020dmfont, park2021lffont}, we use font library labels for style labels $y_s$, and the component labels $U_c$ for content labels $y_c$. The example of component labels is illustrated in Figure~\ref{fig:localized_expert_concept}. The same decomposition rule used by \lffont is adopted. While previous methods only use the style (or content) classifier to train style (or content), we additionally utilize them for the content and style disentanglement by introducing the content-style adversarial loss.

The {\bf generator} $G$ synthesizes a glyph image $\widetilde x$ by combining content and style features from each expert:
\begin{equation}
\label{eq:gen}
    \widetilde x = G ( (f_{s, 1} \circ f_{c, 1}), \ldots, (f_{s, k} \circ f_{c, k})),
\end{equation}
where $\circ$ denotes a concatenate operation.

In the following, we describe the details of each module, training settings, and how to generate samples with only a few references.

\subsection{Learning multiple localized experts with weak local component supervision}
\label{sec:best_allocation}

Our intuition is that extracting different localized features can help each local feature to represent the detailed local structure and fine-grained local style in a complex glyph image. We utilize the compositionality of the font domain to inherit the advantages of component-conditioned methods~\cite{cha2020dmfont, park2021lffont}. Meanwhile, we intentionally remove the explicit component dependency of the feature extractor for achieving generalizability, which is the weakness of previous methods. Here, we employ a multi-headed feature extractor, named {\it multiple localized experts}, where each expert can be specialized for different local concepts. A na\"ive solution is to utilize explicit local supervision, \ie, the pixel-level annotation for each sub-glyph, unable to obtain due to expensive annotation cost. As an alternative, a strong machine annotator can be utilized to obtain local supervision~\cite{yun2021relabel}, but training a strong model, such as the self-trained EfficientNet L2 with 300M images~\cite{xie2020self}, for the font domain is another challenge that is out of our scope.

Utilizing the compositionality, we have the weak component-level labels for the given glyph image, \ie, what components the image has but without the knowledge where they are, similar to the multiple instance learning scenario~\cite{mil, miml}. Then, we let each expert attend on different local concepts by guiding each expert with the component and style classifiers. Ideally, when the number of components $m$ is same as the number of experts, $k$, we expect the $k$ predictions by experts are same as the component labels, and the summation of their prediction confidences is maximized. When $k < m$, we expect the predictions by each expert are ``plausible'' by considering top-k predictions. 

To visualize the role of each expert, we illustrate an example in Figure~\ref{fig:localized_expert_concept}. Presuming three multiple experts, they can learn different local concepts such as the left-side (blue), the right-bottom-side (green), and the right-upper-side (yellow), respectively. Given a glyph composed of four components, the feature from each expert can predict one ($E_1$, $E_2$) or two ($E_3$) labels as shown in the figure. Because we do not want that an expert is explicitly assigned to a component label, \eg, strictly mapping ``人'' component to $E_1$, we solve an automatic allocation algorithm, finding the optimal expert-component matching as shown in Figure~\ref{fig:localized_expert_concept}. Specifically, we formulate the component allocation problem as the Weighted Bipartite B-Matching problem, which can be optimally solved by the Hungarian algorithm~\cite{kuhn1955hungarian}.

From a given glyph image $x$, each expert $E_i$ extracts the content feature $f_{c, i}$. Then, the component feature classifier $Cls_u$ takes $f_{c, i}$ as input and produces the prediction probability $p_i = Cls_u (f_{c, i})$, where $p_i = [ p_{i0}, \ldots , p_{im}]$ and $p_{ij}$ is the confidence scalar value of the component $j$. Let $U_c = \{u_1^c, \ldots, u_m^c\}$ be a set of component labels of the given character $c$, and $m$ be the number of the components. We introduce an allocation variable $w_{ij}$, where $w_{ij}=1$ if the component $j$ is assigned to $E_i$, and $w_{ij} = 0$ otherwise. We optimize the binary variables $w_{ij}$ to maximize the summation over the selected prediction probability such that the number of total allocations is $\max(k, m)$. Now, we formulate the component allocation problem as:
\begin{align}
\begin{split}
\label{eq:allocation_problem}
    &\max_{w_{ij} \in \{0, 1\} | i=1 \ldots k, j \in U_c} \sum_{i=1}^k \sum_{j \in U_c} w_{ij} p_{ij}, \\
    \text{s.t.} \quad &\sum_{i=1}^{k} w_{ij} \geq 1 ~~ \text{for} ~\forall j, \quad \sum_{j \in U_c} w_{ij} \geq 1 ~~ \text{for} ~\forall i, \\
    &\sum_{i=1}^k \sum_{j \in U_c} w_{ij} = \max(k, m), 
\end{split}
\end{align}
where \eqref{eq:allocation_problem} can be reformulated to the Weighted Bipartite B-Matching (WBM) problem, and can be solved by the Hungarian algorithm in a polynomial time $O((m+k)^3)$. We describe the connection between \eqref{eq:allocation_problem} and WBM in the \supp. Now, using the estimated variables $w_{ij}$ in \eqref{eq:allocation_problem}, we optimize auxiliary component classification loss $\mathcal L_{cls, c}$ with the cross entropy loss (CE) as follows:
\begin{equation}
\label{eq:comp_cls_loss}
    \mathcal L_{cls, c, i} (f_{c, i}, U_c) = \sum_{j \in U_c} w_{ij} \text{CE}(Cls_u (f_{c, i}), j).
\end{equation}
Here, we expect that each localized expert is specialized for a specific local concept so that it facilitates the content-style disentanglement. Because the feedback from \eqref{eq:comp_cls_loss} encourages the local features to be better separated into the style and content features, we expect that each expert automatically attends local concepts. We empirically observe that each expert is involved to different local areas without explicit pixel-level supervision (Figure~\ref{fig:expert_cam}).

We additionally formulate the independence between each expert by the Hilbert-Schmidt Independence Criterion~\cite{HSIC} which has been used in practice for statistical testing~\cite{HSIC,HSICTest}, feature similarity measurement~\cite{kornblith2019similarity}, and model regularization~\cite{quadrianto2019discovering,zhang2018ijcai, bahng2019rebias}. HSIC is zero if and only if two inputs are independent of each other. Since HSIC is non-negative, the independence criterion can be achieved by minimizing HSIC. Under this regime, we use HSIC and lead the local feature $f_i$ extracted by $E_i$ independent to the other local features $f_{i^\prime}$ as follows:
\begin{equation}
\label{eq:loss_indp_exp}
    \mathcal L_{\text{indp exp}, i} = \sum_{i^\prime=1, i^\prime \neq i}^k \text{HSIC}(f_{i}, f_{i^\prime}).
\end{equation}
We leave the detailed HSIC formulation is in the \supp.

\subsection{Content and style disentanglement}
\label{sec:cs_disentangle}
To achieve perfect content and style disentanglement, the style (or content) features should include the style (or content) domain information but exclude the content (or style) domain information. We employ two objective functions for this: {\it content-style adversarial loss} and {\it independent loss}.

The {\bf content-style adversarial loss}, motivated by the domain adversarial network~\cite{ganin2016domain}, enforces the extracted features for style (or content) is useless to classify content (or style). Thus, a style feature $f_{s, i}$ is trained to satisfy (1) correctly classify a style label $y_s$ by the style classifier $Cls_s$ with the cross entropy loss (CE) and (2) fooling the content labels predicted by the component classifier $Cls_u$. Specifically, we maximize the entropy ($H$) of the predicted probability to enforce the uniform prediction. Formally, we define our objective function for a style feature $f_{s, i}$ as follows:
\begin{equation}
\label{eq:loss_cls_style}
    \mathcal{L}_{s, i} (f_{s, i}, y_s) = \text{CE}(Cls_s(f_{s, i}), y_s) - H(Cls_u(f_{s, i})).
\end{equation}

We define $\mathcal{L}_{c, i}$ as the objective function for a content feature $f_{c, i}$ employs $\mathcal{L}_{cls,c,i}$ \eqref{eq:comp_cls_loss} instead of the cross entropy of $y_c$ as follows:
\begin{equation}
\label{eq:loss_cls_content}
    \mathcal{L}_{c, i} (f_{c, i}, U_c) = \mathcal L_{cls, c, i}(f_{c, i}, U_c) - H(Cls_s(f_{c, i})).
\end{equation}

We also employ the independence loss between content and style local features, $f_{c, i}$ and $f_{s, i}$ for the disentanglement of content and style representations. That is:
\begin{equation}
\label{eq:loss_indp_factor}
    \mathcal L_{\text{indp}, i} = \text{HSIC}(f_{s, i}, f_{c, i}).
\end{equation}

\subsection{Training}
We train our model to synthesize a glyph image from the given content and style labels using the Chinese font dataset (details in \S\ref{sec:dataset}). More specifically, we construct a mini-batch, where $n$ glyphs share the same content label $y_c$ (from random styles), and $n$ glyphs share the same style label $y_s$ (from random contents). Then, we let the model generate a glyph with the content label $y_c$ and the style label $y_s$. In our experiments, we set $n=3$ and synthesize 8 different glyphs in parallel, \ie, the mini-batch size is 24.

We employ a discriminator module $D$ and the generative adversarial loss~\cite{gan} to achieve high-quality visual samples. In particular, we use the hinge generative adversarial loss $\mathcal L_{adv}$~\cite{zhang2019sagan}, feature matching loss $\mathcal L_{fm}$, and pixel-level reconstruction loss $\mathcal L_{recon}$ by following the previous high fidelity GANs, \eg, BigGAN~\cite{brock2018large}, and state-of-the-art font generation methods, \eg, DM-Font~\cite{cha2020dmfont} or LF-Font~\cite{park2021lffont}. The details of each objective function are in the \supp.

Now we describe our full objective function. The entire model is trained in an end-to-end manner with the weighted sum of all losses, including  \eqref{eq:loss_indp_exp}, \eqref{eq:loss_cls_style}, \eqref{eq:loss_cls_content}, and \eqref{eq:loss_indp_factor}.
\begin{align}
\begin{split}
    \mathcal L_{D} &= \mathcal L_{adv}^D, \\
    \mathcal L_{G} &= \mathcal L_{adv}^G + \lambda_{recon} \mathcal L_{recon} + \mathcal L_{fm}\\
    \mathcal L_{exp} &= \sum_{i=1}^k \left [ \mathcal L_{s, i} + \mathcal L_{c, i} + \mathcal L_{\text{indp}, i} + \mathcal L_{\text{indp exp}, i} \right ]
\end{split}
\end{align}
As conventional GAN training, we alternatively update $\mathcal L_{D}$, $\mathcal L_{G}$, and $\mathcal L_{exp}$. The control parameter $\lambda_{recon}$ is set to 0.1 in our experiments. We use Adam optimizer~\cite{adam}, and run the optimizer for 650k iterations. We additionally provide the detailed training setting is in the \supp.

\subsection{Few-shot generation}
\label{sec:generation}

When the source and a few reference glyphs are given, \ours extract the content features from the source glyphs and the style features from the reference glyphs. Assume we have $n_r$ number of reference glyphs $x^r_1, \ldots, x^r_{n_r}$ with a coherent style $y_{s^r}$. First, our multiple experts $\{E_1, \ldots, E_k\}$ extract localized style feature $[f_{s^r, i}^1, \ldots, f_{s^r, i}^{n^r}]$ for $i=1\ldots k$ from the reference glyphs. Then, we take an average over the localized features to represent a style representation, \ie, $f_{s^r, i} = \frac{1}{n^r}\sum_{j=1}^{n^r} f_{s^r, i}^j$ for $i=1\ldots k$. Finally, the style representation is combined with the content representation extracted from the known source glyph to generate unseen style glyph.

\section{Experiments}

In this section, we describe the evaluation protocols, and experimental settings. We extend previous \ffg benchmarks to unseen language domain to measure the generalizability of a model. \ours is compared with four \ffg methods on the proposed extended \ffg benchmark via both the qualitative and quantitative evaluations. Experimental results demonstrate that \ours outperforms existing methods in the most of evaluation metrics. The ablation and analysis study helps understand the role and effects of our multiple experts and objective functions.

\subsection{Comparison methods}

\noindent\textbf{Universial style representation methods.}
{\bf EMD}~\cite{zhang2018_cvpr_emd} adopts content and style encoders that extract universal content and style features from a few reference glyphs. {\bf AGIS-Net}~\cite{gao2019agisnet} proposes the local texture refinement loss to handle unbalance between the number of positive and negative samples. {\bf FUNIT}~\cite{liu2019funit} is not directly proposed for \ffg task, but we employ the modified version of FUNIT as our comparison method following previous works~\cite{cha2020dmfont, park2021lffont}.

\noindent\textbf{Component-conditioned methods.}
{\bf DM-Font}~\cite{cha2020dmfont} learns two embedding codebooks (or the dual-memory) conditioned by explicit component labels. When the target character contains a component either unseen during training or not in the reference set, DM-Font is unable to generate a glyph. As these drawbacks are impossible to be fixed with only minor modifications, we do not compate DM-Font to \ours. {\bf LF-Font}~\cite{park2021lffont} relaxes the restriction of DM-Font by estimating missing component features via factorization module. Although LF-Font is still not applicable to generate a character with unseen components, we slightly modify LF-Font (as described in the \supp) and compare the modified version with other methods.

\subsection{Evaluation protocols}
\label{sec:dataset}

To show the generalizability to the unseen language systems, we propose an extended \ffg scenario; training a \ffg model on a language system and evaluating the model on the other language system. In this paper, we first train \ffg models on the Chinese font dataset, and evaluate them on both Chinese generation ({\it in-domain transfer scenario}) and Korean generation ({\it zero-shot cross-lingual scenario}).

\noindent\textbf{Dataset.}
We use the same Chinese font dataset collected by Park~\etal~\cite{park2021lffont} for training. The dataset contains 3.1M Chinese glyph images with 467 different styles, and 19,514 characters are covered. We also use the same decomposition rule as Park~\etal~\cite{park2021lffont} to extract component labels. We exclude 28 fonts, and 214 Chinese characters from the training set, and use them to evaluation. For the Korean \ffg evaluation, we use the same test characters with Cha~\etal~\cite{cha2020dmfont}, 245 characters. To sum up, we evaluate the methods by using 28 font styles with 214 Chinese and 245 Korean characters.

\input{tables/main_table}

\noindent\textbf{Evaluation metrics.}
Due to the style of the font domain is defined by a local fine-grained shape, \eg, stroke, size, or serif-ness, measuring the visual quality with a unified metric is a challenging problem. A typical challenge is the multiplicity of the font styles; because the font style is defined locally, there could be multiple plausible glyphs satisfying our objectives. However, we only have one ``ground truth'' glyphs in the test dataset. Furthremore, for the Korean generation task with Chinese references, we even do not have ``ground truth'' Korean glyphs with the reference styles. Thus, we need to employ evaluation metrics that does not require ground truth, and can evaluate plausibility of the given samples. We therefore use four different evaluation metrics to measure the visual quality in various viewpoints.

Following previous works~\cite{cha2020dmfont, park2021lffont}, we train evaluation classifiers that classifies character labels (content-aware) and font labels (style-aware). Note that these classifiers are only used for evaluation, and trained separately to the \ffg models. We train three classifiers, the style classifier on the Chinese test fonts, the content classifier on the Chinese test characters, and the content classifier on the Korean test characters. The details of the evaluation classifiers are in the \supp. Using the classifiers, we measure the {\bf classification accuracies} for style and content labels. We also report the accuracy when both classifiers are correctly predicted.

We conduct a {\bf user study} for quantifying the subjective quality. The participants are asked to pick the three best results, considering the style, the content, and the most preferred considering both the style and the content. All 28 test styles with 10 characters are shown to the participants. For each test style, we show Chinese and Korean samples separately to the users. \Ie, a participant picks $28 \times 3 \times 2 = 168$ results. We collect the responses from 57 participants. User study samples are in the \supp.

We also report LPIPS~\cite{zhang2018_cvpr_lpips} scores to measure the dissimilarity between the generated images and their corresponding ground truth images, thus it is only reported for Chinese \ffg task. Using the style and content classifiers, Frechét inception distance (FID) ~\cite{heusel2017_nips_ttur_fid} between the generated images and real images are computed and their harmonic mean is reported (FID(H)). We describe the details in the \supp.

\subsection{Experimental results}\label{sec:eval}

\noindent\textbf{Quantitative evaluation.}
Table~\ref{table:main_fewshot} shows the \ffg performances by \ours and competitors. The reported values are the average of 50 different experiments, where four reference images per style are used for font generation in each experiment. In the table, we observe that \ours outperforms other methods in the both in-domain transfer scenario and zero-shot cross-lingual generation scenario with the most of evaluation metrics. Especially, \ours remarkably outperforms other methods in the cross-lingual task. In the in-domain transfer scenario, ours exceeds others in the classification accuracies and the user study. We observe that \ours perform worse than others in the Chinese FID, where FID is known to sensitive to noisy or blur images, regardless of the image quality itself~\cite{ramesh2021zero}. Our method shows the remarkably better performances in more reliable evaluation, user study in all criterions.

\noindent\textbf{Qualitative evaluation.}
We illustrated the generated samples in Figure~\ref{fig:main_few}. We show four reference images to extract each style in the top row, and the source images in the second row where each source image is used to extract the content. In the green box in Figure~\ref{fig:main_few}, we observe that AGIS-Net often fails to reflect the reference style precisely and generate local details. FUNIT generally shows similar trends with AGIS-Net, while FUNIT often produces shattered glyphs when the target glyph and the source glyph have significantly different structures (red box). At a glance, LF-Font seems to capture the detailed local styles well. However, it often misses important detailed local component such as dot and stroke, as shown in the yellow box. Comparing to other methods, \ours synthesizes the better detailed structures both in content and style, owing to the strong representation power of locally specialized experts. The advantage of \ours is highlighted in the cross-lingual \ffg. All existing models often generate unrecognizable characters under the cross-lingual scenario. Nevertheless, \ours preserves both the detailed local style and content and generates the plausible and recognizable images consistently. Such a noticeable gap in visual quality explains the large performance leap of \ours in the user study. 

\begin{figure}
    \centering
    \includegraphics[width=0.16\linewidth]{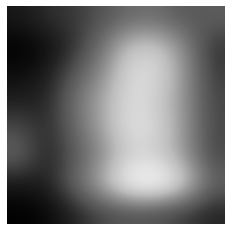}%
    \includegraphics[width=0.16\linewidth]{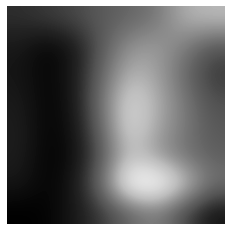}%
    \includegraphics[width=0.16\linewidth]{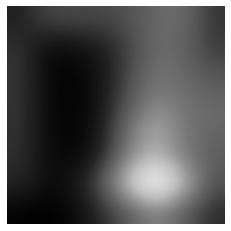}%
    \includegraphics[width=0.16\linewidth]{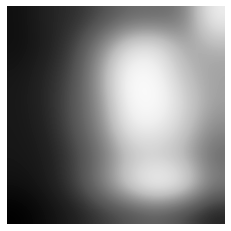}%
    \includegraphics[width=0.16\linewidth]{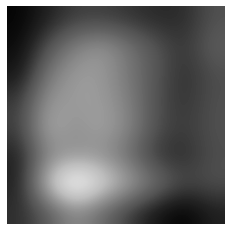}%
    \includegraphics[width=0.16\linewidth]{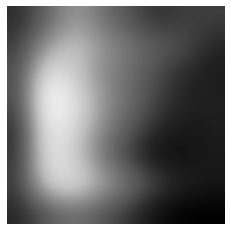}%
    \caption{\small {\bf Each localized expert attends different local areas.} We show the variance of Class Activation Maps (CAMs) on training images for each expert. The brighter intensity indicates that the variance of CAMs is higher in that region. }
    \label{fig:expert_cam}
    \vspace{-.5em}
\end{figure}

\begin{figure}[t]
    \centering
    \includegraphics[width=\linewidth]{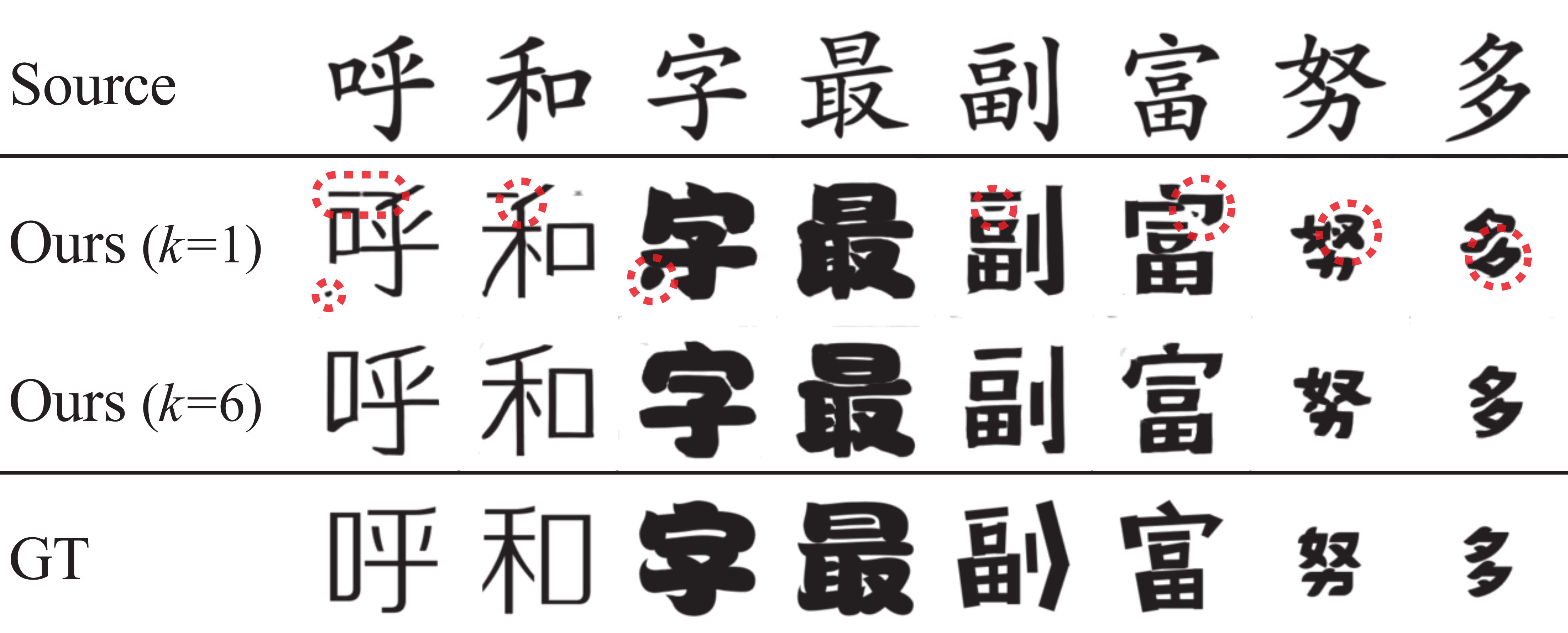}%
    \caption{\small {\bf Generated samples of the models having different number of heads.} The samples generated with four reference glyphs by the single-headed model and multi-headed model are shown. We highlight the defects in red dotted circles that appeared in the images generated by the single-expert model. $k$ denotes the number of experts.}%
    \label{fig:num_head}
\end{figure}

\begin{table}[t]
\small
\centering
\begin{tabular}{@{}ccccc@{}}
\toprule
  & Acc (S) $\uparrow$ & Acc (C) $\uparrow$ & Acc (B) $\uparrow$ & LPIPS $\downarrow$\\
\midrule
 Ours ($k=1$) &  72.2 & 98.7 & 71.4 & 0.133\\
 Ours ($k=6$) &  78.9 & 99.5 & 78.7 & 0.120\\
\bottomrule
\end{tabular}
\vspace{.5em}
\caption{\small {\bf Impact of the number of experts $k$.}
Single-expert model ($k=1$) and multiple-experts model ($k=6$, proposed) are compared on in-domain Chinese transfer benchmark.}
\label{table:abl_exp} 
\end{table}

\subsection{Analyses}

\noindent\textbf{Learned local concepts by different experts.}
We show the local concepts learned by each expert by visualizing where each expert attends on. We extract the Class Activation Maps (CAMs) of the training samples using the component classifier $Cls_u$ on each local feature. Then, we visualize the variance of CAMs in Figure~\ref{fig:expert_cam}. In Figure~\ref{fig:expert_cam}, the region of each image with bright intensity than the surrounding indicates the region where each expert pays more attention. Interestingly, without any explicit pixel-level annotation, our localized experts attend different local areas of the images. These maps support that each expert of \ours tends to cover different local areas of the input image. Summarizing, these experimental studies demonstrate that {\it multiple localized experts} capture different local areas of the input image as we intend, and employing {\it multiple localized experts} helps us to enhance the quality of generated images by preserving the local details during the style-content disentanglement.

\noindent\textbf{Multiple experts vs. single expert.}
We compare the performances of the single expert model ($k=1$) with our multiple expert model ($k=6$) on benchmark in-domain transfer scenario. The results are shown in Table~\ref{table:abl_exp} and Fig~\ref{fig:num_head}. We observe that using multiple heads is better than a single head in the classification accuracies. We also observe that the generated images by the single-headed model fails to preserve the local structures delicately, \eg important strokes are missing, while the multi-headed model captures local details well.

\noindent\textbf{Character labels vs. local component labels.} We assume that component supervision is beneficial to learn experts with different local concepts. We replace the component supervision (multiple image-level sub-concepts) to the character supervision (single image-level label). Table~\ref{table:abl_charAC} shows that utilizing character supervision incurs a mode collapse. We speculate that two reasons caused the collapse, (1) the number of characters ($\approx 19k$) is too large to learn, while the number of components is reasonably small ($371$), and (2) our best allocation problem prevents the experts from collapsing into the same values, while the character supervised model has no restriction to learn different concepts.

\begin{table}[t]
\small
\centering
\begin{tabular}{@{}ccccc@{}}
\toprule
  & Acc (S) $\uparrow$ & Acc (C) $\uparrow$ & Acc (B) $\uparrow$ & LPIPS $\downarrow$\\
\midrule
 Ours ($Cls_u$) &  78.9 & 99.5 & 78.7 & 0.120\\
 Ours ($Cls_c$) &  94.8 & 0.04 & 0.04 & 0.214\\
\bottomrule
\end{tabular}
\vspace{.5em}
\caption{\small {\bf Comparing the component classifier and the character classifier as weak supervision.} We compare two auxiliary classifiers as content supervision.
Ours ($Cls_u$) denotes \ours using the component classifier and Ours ($Cls_c$) denotes the model replaced the component classifier to the character classifier.}
\vspace{-.5em}
\label{table:abl_charAC} 
\end{table}

\begin{table}[t]
\small
\centering
\begin{tabular}{@{}cccccc@{}}
\toprule
$\mathcal{L}_{indp, i}$ & $\mathcal{H}_{c,s}$ & $\mathcal{L}_{c,s}$ & Acc (S) & Acc (C) & Acc (B) \\
\midrule
\yesmark & \yesmark & \yesmark &  \textbf{59.0} & \textbf{95.9} & \textbf{56.8} \\
\nomark & \yesmark & \yesmark &  52.0 & 95.8 & 50.0 \\
\nomark & \nomark & \yesmark &  51.6 & 95.5 & 49.4 \\
\nomark & \nomark & \nomark & 27.8 & 89.1 & 24.7 \\
\midrule
\multicolumn{3}{c}{\lffont} & 38.5 & 95.2 & 36.5 \\
\bottomrule
\end{tabular}
\vspace{.5em}
\caption{\small {\bf Impact of loss functions.} We compare models by ablating the proposed object functions trained and tested on Korean-handwriting dataset. The results show that the content-style adversarial loss $\mathcal{L}_{c,s}$ and the maximizing entropy term $\mathcal{H}_{c,s}$ and independent loss $\mathcal{L}_{indp, i}$ are all important components.}
\label{table:abl_loss}
\vspace{-1em}
\end{table}

\noindent\textbf{Loss ablations.}
We investigate the effect of our loss function design by the models trained and tested on Korean handwritten dataset. The evaluation results are reported in Table~\ref{table:abl_loss}. The detailed training settings are in \supp. $\mathcal{H}_{c,s}$ denotes the maximizing entropy terms in the content-style adversarial loss, and $\mathcal{L}_{c,s}$ denotes the content-style adversarial loss. Table~\ref{table:abl_loss} shows that all the proposed loss functions for the style-content disentanglement are effective to enhance the overall performances.

\section{Conclusion}

We propose a novel few-shot font generation method, namely \ours. Our goal is to achieve both the rich representation for the local details and the generalizability to the unseen component and language. To this end, \ours employ {\it multi-headed encoder}, trained by weak local component supervision, \ie style and content feature classifiers. Based on interactions between these {\it feature classifiers} and localized experts, \ours learns to disentangle the style and content successfully by developing localized features. Finally, the proposed model generates the plausible font images, which preserve both local detailed style of the reference images and precise characters of the source images. Experimental results show that \ours outperforms existing methods in in-domain transfer scenario and zero-shot cross-lingual transfer scenario; especially large performance leap in the cross-lingual scenario.

\section*{Acknowledgements}
All experiments were conducted on NAVER Smart Machine Learning (NSML)~\cite{nsml} platform.

{\small
\bibliographystyle{ieee_fullname}
\bibliography{egbib}
}

\appendix

\numberwithin{equation}{section}
\numberwithin{figure}{section}
\numberwithin{table}{section}

\section*{Appendix}

We describe additional experimental results to complement the main paper (\S\ref{sec:appendix-exp}). The implementation details are in \S\ref{sec:appendix-training}. Finally, we provide the detailed evaluation protocols (\S\ref{sec:appendix-eval}).

\section{Additional experimental results}
\label{sec:appendix-exp}

\subsection{More visual examples}

\begin{figure*}[t]
    \centering
    \includegraphics[width=\linewidth]{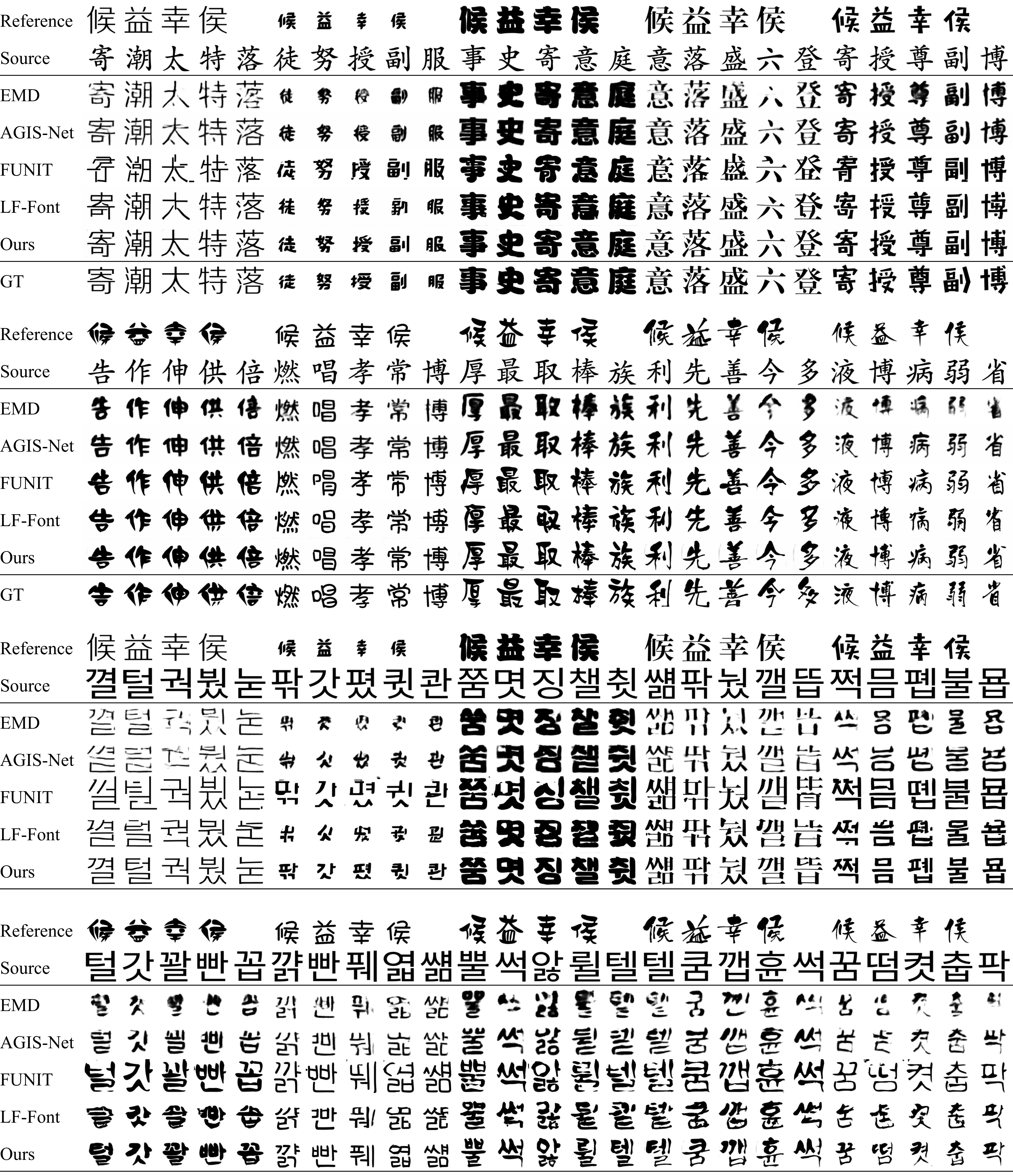}%
    \caption{\small {\bf Generation samples.} We provide more generated glyphs with four reference glyphs.}%
    \label{fig:results}
    \vspace{3em}
\end{figure*}

We show more generated glyphs in Figure~\ref{fig:results}. \ours correctly synthesizes the strokes, dot, thickness and size of the ground truth glyphs. In the cross-lingual \ffg, \ours can produce promising results in that they are all readable. Meanwhile, all other competitors provide inconsistent results, which are often impossible to understand. These results show a similar conclusion as our main paper.

\subsection{Impact of the number of experts}

\begin{figure}[t]
    \centering
    \includegraphics[width=\linewidth]{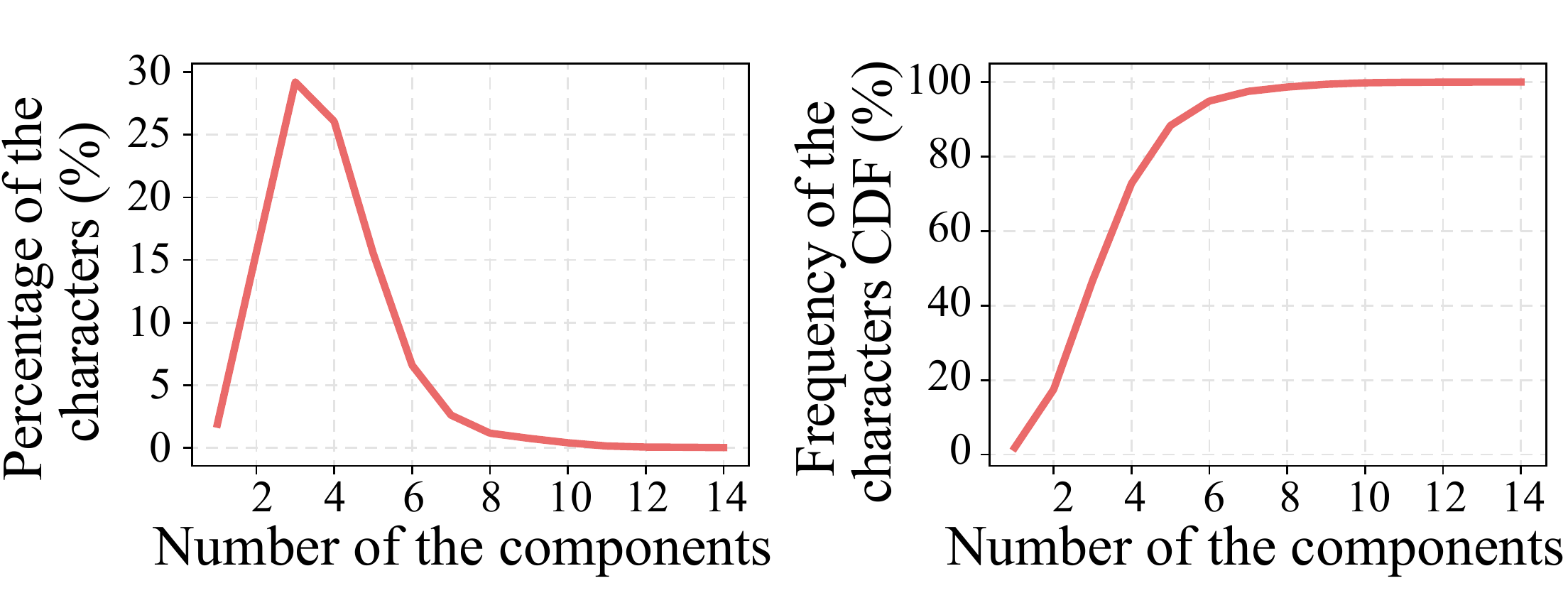}%
    \caption{\small {\bf The distribution of number of components.} The left shows the percentage of characters with different number of components and the right shows the cumulative summation of the left.}
    \label{fig:num_comps}
\end{figure}

In Table~\ref{table:abl_exp_k}, we report the performances by varying the number of experts, $k$. We observe that larger $k$ brings better performances until $k=6$, but larger $k$, \eg, 8, shows slightly worse performance than $k=6$. We presume that this is because there are no sufficient data having more than or equal to eight components for training all the eight experts to capture different concepts. Figure~\ref{fig:num_comps} illustrates the frequency of the number of components. From this graph, we find that the most characters have less than 8 components in our Chinese dataset. Moreover, larger $k$ means the number of parameters are increased, resulting in more training and inference runtime. Hence, in the paper, we choose $k=6$ for all experiments.

\begin{table}[h]
\small
\centering
\begin{tabular}{@{}ccccc@{}}
\toprule
$k$  & Acc (S) $\uparrow$ & Acc (C) $\uparrow$ & Acc (B) $\uparrow$ & LPIPS $\downarrow$\\
\midrule
1 &  72.2 & 98.7 & 71.4 & 0.133\\
2 &  79.0 & 99.3 & 78.5 & 0.128\\
6 &  78.9 & 99.5 & 78.7 & 0.120\\
8 &  75.5 & 99.5 & 75.2 & 0.123\\
\bottomrule
\end{tabular}
\vspace{.5em}
\caption{\small {\bf Impact of the number of experts $k$.}
The models with different number of heads are compared on in-domain Chinese transfer benchmark.
}
\label{table:abl_exp_k} 
\end{table}

\section{Implementation details} 
\label{sec:appendix-training}
\subsection{Network architecture}
Each localized expert $E_i$ has 11 layers including convolution, residual, global-context~\cite{cao2019gcnet}, and convolutional block attention (CBAM)~\cite{woo2018cbam} blocks. The multiple localized experts share the weights of their first five blocks. The two feature classifiers $Cls_s$ and $Cls_u$ have the same structure; a linear block following two residual blocks. The weights of the first two residual blocks are shared. The generator $G$ consists of convolution and residual blocks. Please refer our code for the detailed architecture.

\subsection{Component allocation problem to weighted bipartite B-matching problem}
Given a bipartite graph $G = (V, E)$, where $V$ is a set of vertices, $E$ is a set of edges and $W$ is the weight values for each edge $e \in E$, the weighted bipartite B-matching (WBM) problem~\cite{kleinschmidt1995strongly} aims to find subgraph $H = (V, E^\prime)$ maximizing $\sum_{e \in E} W(e)$ with every vertex $v \in V$ adjacent to at most the given budget, $B(v)$, edges. WBM problem can be solved by the Hungarian algorithm~\cite{kuhn1955hungarian}, a typical algorithm to solve combinatorial optimization in a polynomial time, in $O(|V| |E|) = O(|V|^3)$. For curious readers, we refer recent papers solving variants of WBM problems~\cite{chen2016conflict, ahmed2017diverse}.

We recall the component allocation problem described in the main paper:
\vspace{-.5em}
\begin{align}
\begin{split}
\label{eq:allocation_problem_appendix2}
&\max_{w_{ij} \in \{0, 1\} | i=1 \ldots k, j \in U_c} \sum_{i=1}^k \sum_{j \in U_c} w_{ij} p_{ij}, \\
    \text{s.t.} \quad &\sum_{i=1}^{k} w_{ij} \geq 1 ~~ \text{for} ~\forall j, \quad \sum_{j \in U_c} w_{ij} \geq 1 ~~ \text{for} ~\forall i, \\
&\sum_{j \in U_c} w_{ij} \leq \max\left(1, \ceil[\bigg]{\frac{m}{k}}\right) ~\text{for}~ \forall i \\
&\sum_{i=1}^k w_{ij} \leq \max\left(1, \ceil[\bigg]{\frac{k}{m}}\right) ~\text{for}~ \forall j. 
\end{split}
\end{align}

We replace the last condition, $\sum_{i=1}^k \sum_{j \in U_c} w_{ij} = \max(k, m)$ to the upper bound condition where $\ceil[]{\cdot}$ denotes the ceiling function. For example, if $k = 3$ and $m=4$, the budget for each expert is $2$, while the budget for each component is $1$. We build a bipartite graph where the vertex set contains all experts and all valid components, and the edge weights are the prediction probability $p_{ij}$. Now \eqref{eq:allocation_problem_appendix2} can be re-formulated by the WBM problem.

\subsection{HSIC Formulation}

When training \ours, we let the two feature outputs from different experts, or content and style features independent of each other. To measure the independence between content feature and style feature, we first assume that the content features $f_c$ and the style features $f_s$ are drawn from two different random variables, $Z_c$ and $Z_s$, \ie, $f_c \sim Z_c$ and $f_s \sim Z_s$. We employ Hilbert Schmidt independence criterion (HSIC)~\cite{HSIC} to measure the independence between two random variables. For two random variables $Z_c$ and $Z_s$, HSIC is defined as $\text{HSIC}^{k,l}(Z_c,Z_s):=||C^{k,l}_{Z_c Z_s}||_{\text{HS}}^2$ where $k$ and $l$ are kernels, $C^{k,l}$ is the cross-covariance operator in the Reproducing Kernel Hilbert Spaces (RKHS) of $k$ and $l$, $||\cdot ||_{\text{HS}}$ is the Hilbert-Schmidt norm~\cite{HSIC, HSICTest}. If we use radial basis function (RBF) kernels for $k$ and $l$, HSIC is zero if and only if two random variables are independent.

Since we only have the finite number of samples drawn from the distributions, we need a finite sample estimator of HSIC. Following Bahng~\etal~\cite{bahng2019rebias}, we employ an unbiased estimator of HSIC, $\text{HSIC}^{k,l}_1(Z_c,Z_s)$~\cite{unbiasedHSIC} with $m$ samples. Formally, $\text{HSIC}^{k,l}_1(Z_c,Z_s)$ is defined as:
\begin{align}
\begin{split}
\label{eq:unbiased_hsic}
    \text{HSIC}^{k,l}_1(Z_c,Z_s) =\frac{1}{m(m-3)}\bigg[\text{tr}(\widetilde{Z}_c\widetilde{Z}_s^T)     ~+ \\\frac{\mathbf{1}^T \widetilde{Z}_c \mathbf{1} \mathbf{1}^T \widetilde{Z}_s^T \mathbf{1}}{(m-1)(m-2)}
    - \frac{2}{m-2} \mathbf{1}^T \widetilde{Z}_c \widetilde{Z}_s^T  \mathbf{1} \bigg] 
\end{split}
\end{align}
where ($i$, $j$)-th element of a kernel matrix $\widetilde{Z}_{c}$ is defined as, $\widetilde{Z}_{c}(i, j) = (1-\delta_{ij})\,k(f_{c}^i,f_{c}^j)$, and the $i$-th feature in the mini-batch $f_c^i$, is assumed to be sampled from the $Z_c$, \ie, $\{f_c^i\}\sim Z_c$. We similarly define $\widetilde{Z}_{s}(i, j) = (1-\delta_{ij})\,l(f_{s}^i,f_{s}^j)$.

In practice, we compute $\text{HSIC}^{k,l}_1(Z_c,Z_s)$ in a mini-batch, \ie, $m$ is the batch size. We use the RBF kernel with kernel radius $0.5$, \ie, $k(f_{c}^i,f_{c}^j) = \exp(- \frac{1}{2} \| f_{c}^i - f_{c}^j \|_2^2)$.

\subsection{GAN objective details}
We employ two conditional discriminators $D_s$ and $D_c$ which predict a style label $y_s$ and a content label $y_c$, respectively. In practice, we employ a multitask discriminator $D$, and different projection embeddings for content labels and style labels, following the previous methods \cite{liu2019funit, cha2020dmfont, park2021lffont}. The hinge loss \cite{zhang2019sagan} is employed to high fidelity generation:
\begin{align}
\begin{split}
    \mathcal L_{adv}^D =~~~~ &\mathbb E_{(x, y_c, y_s)} \left[ [1 - D(x, y_s) ]_+ + [1 - D(x, y_c) ]_+ \right] \\
    + &\mathbb E_{(\widetilde x, y_c, y_s)} \left[ [1 - D(\widetilde x, y_s) ]_+ + [1 - D(\widetilde x, y_c) ]_+ \right] \\
    \mathcal L_{adv}^G = -& \mathbb E_{(\widetilde x, y_c, y_s)} \left[ D(\widetilde x, y_s) + D(\widetilde x, y_c) \right],
\end{split}
\end{align}
where $\widetilde x$ is the generated image by combining a content feature extracted from an image with content label $y_c$ and a style feature extracted from an image with style label $y_s$.

The feature matching loss $\mathcal{L}_{fm}$ and the reconstruction loss $\mathcal{L}_{recon}$ are formulated as follows:
\begin{align}
\begin{split}
    &\mathcal L_{fm} ~~~ = ~\mathbb E_{(x, \widetilde x)} \left[\sum_{l=1}^{L-1} \| D^{l}(x) - D^{l}(\widetilde x) \|_1 \right], \\
    &\mathcal L_{recon} = ~\mathbb E_{(x, \widetilde x)} \left[ \| x - \widetilde x \|_1 \right],
\end{split}
\end{align}
where $L$ is the number of layers in the discriminator $D$ and $D^l$ denotes the output of $l$-th layer of $D$.

\subsection{Training details}
We use Adam \cite{adam} optimizer to optimize the \ours. The learning rate is set to 0.001 for the discriminator and 0.0002 for the remaining modules. The mini-batch is constructed with the target glyph, style glyphs, and content glyph during training. Specifically, we first pick the target glyph randomly. Then, we randomly select $n$ style glyphs with the same style as the target glyph, and $n$ content glyphs with the same character as the target glyph for each target glyph. Here, the target glyph is excluded from the style and content glyphs selection. We set $n$ to $3$ during training. We set the number of heads $k$ to 6 and train the model for 650k iteration with the full objective functions for the Chinese glyph generation. For the Korean, we set the number of heads $k$ to 3 and train the model for 200k iteration with the all objective functions except $\mathcal{L}_{\text{indp exp,}i}$. We do not employ the $\mathcal{L}_{\text{indp exp,}i}$ during training for the Korean glyph generation, due to the special characteristic of the Korean script; always decomposed to fixed number of components, \eg, 3.

\section{Evaluation details}
\label{sec:appendix-eval}

\subsection{Classifiers}
Three classifiers are trained for the training; the style classifier, the Chinese character classifier, and the Korean character classifier. The style classifier and the Chinese character classifier are trained with the same Chinese dataset, including 209 Chinese fonts and 6428 Chinese characters per font. Besides, we used the Korean dataset that DM-Font~\cite{cha2020dmfont} provides to train the Korean character classifier. The classifiers have ResNet-50~\cite{he2016_cvpr_resnet} structure. We optimize the classifiers using AdamP optimizer~\cite{heo2021adamp} with learning rate 0.0002 for 20 epochs. During training, the CutMix augmentation~\cite{yun2019cutmix} is adopted and the mini-batch size is set to 64.

\subsection{LF-Font modification}
Since LF-Font~\cite{park2021lffont} cannot handle the unseen components in the test time due to its component-conditioned structure, we modify its structure to enable the cross-lingual font generation. We loose the component-condition of LF-Font in the test time only, by skipping the component-condition block when the unseen component is given. Note that, we use original LF-Font structure for the training to reproduce its original performance.

\subsection{User study examples}
\begin{figure}[t]
    \centering
    \includegraphics[width=\linewidth]{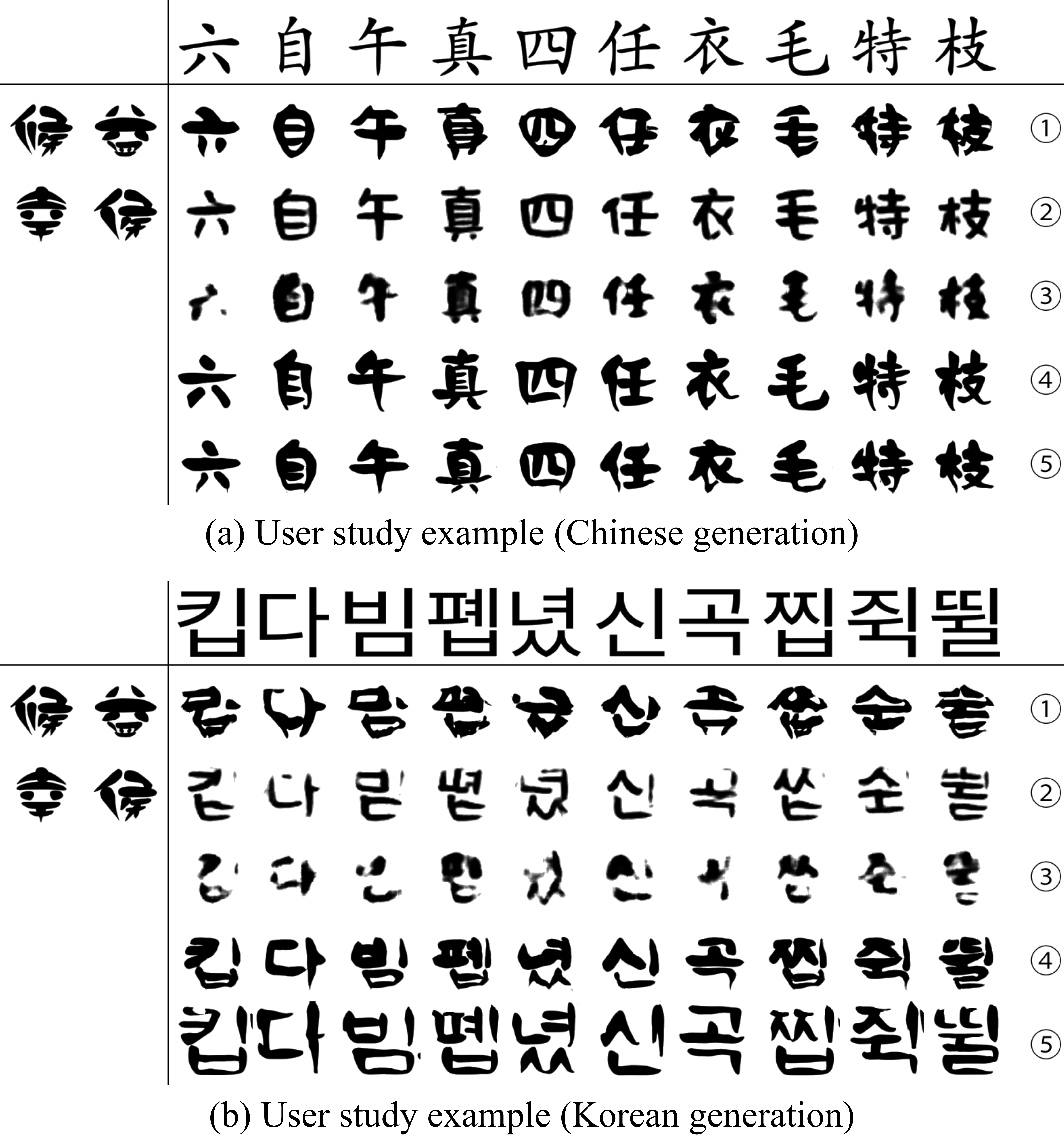}%
    \caption{\small {\bf User study examples.} The example images that we provide to the candidates are shown. Each image includes the reference images, source images, and the generated images.}%
    \label{fig:userstudy}
\end{figure}

We show the sample images used for the user study in Figure~\ref{fig:userstudy}. Five methods, including EMD~\cite{zhang2018_cvpr_emd}, \agis, \funit, \lffont, and \ours are randomly displayed to users for every query.

\subsection{FID}
We measure the style-aware and content-aware Frechét inception distance (FID) ~\cite{heusel2017_nips_ttur_fid} between generated images and rendered images using the style and content classifier. For the Chinese glyphs, the style-aware and content-aware FIDs are measured with the generated glyphs and the corresponding ground truth glyphs. Since the ground truth glyphs of cross-lingual generation do not exist, the style-aware FID is measured the generated glyphs and all the available rendered glyphs having the same style with the generated images. The content-ware FID is measured similar to the style-aware FID. The style-aware (S) and the content-aware (C) FID values and their harmonic mean (H) are reported in Table~\ref{table:FID}. Despite that \ours shows the slight degradation in FID for Chinese font generation, these results are not consistent with the user study and qualitative evaluation. For quantifying the image quality, we tend to trust the user study more because it better reveals the user's preference. 

\begin{table}[t]
\small
\centering
\setlength{\tabcolsep}{5pt}
\begin{tabular}{@{}lccccccc@{}}
\toprule
 & \multicolumn{3}{c}{CN $\xrightarrow{}$ CN} && \multicolumn{3}{c}{CN $\xrightarrow{}$ KR} \\ 
FIDs & \multicolumn{1}{c}{S} & \multicolumn{1}{c}{C} & \multicolumn{1}{c}{H} && \multicolumn{1}{c}{S} & \multicolumn{1}{c}{C} & \multicolumn{1}{c}{H} \\ \midrule
EMD         & 145.5 & 51.1 & 79.7 && 220.3 & 113.8 & 150.0 \\
AGIS-Net    & 91.0  & 10.8 & 19.2 && 235.5 & 106.5 & 146.5 \\
FUNIT       & 50.6  & 11.8 & 19.2 && 486.4 & 107.4 & 176.0 \\
LF-Font     & \textbf{43.5}  & \textbf{9.0}  & \textbf{14.8} && 187.8 & 123.4 & 148.7 \\
\ours       & 50.5  & 13.9 & 21.8 && \textbf{113.2} & \textbf{78.1}  & \textbf{84.1} \\ \bottomrule
\end{tabular}
\vspace{0.5em}
\caption{We provide style-aware (S), content-aware(C) FIDs measured by the style and content classifiers. The harmonic mean (H) of the style-aware and the content-aware FIDs values are identical to the values reported in the main table.
}
\label{table:FID} 
\end{table}

\end{document}

%% file: tables/main_table.tex
\begin{table*}[ht!]
\centering
\small
\setlength{\tabcolsep}{4pt}
\begin{tabular}{@{}cclccccccccccc@{}}
\toprule
                  &&                   & Acc (S) \% & Acc (C) \% & Acc (B) \% &  &  User (S) \%&  User (C) \%&  User (B) \%& & LPIPS $\downarrow$ & FID (H) $\downarrow$ \\ \midrule
\parbox[t]{0mm}{\multirow{5}{*}{\rotatebox[origin=c]{90}{CN $\xrightarrow{}$ CN}}}   
                  && EMD (CVPR'18)     & 6.6 & 51.3 & 4.6 && 0.7 & 0.1 & 0.3 && 0.212 & 79.7       \\
                  && AGIS-Net (TOG'19) & 25.5 & \textbf{99.5} & 25.4 && 22.4 & \textbf{34.2} & 26.8 && 0.124 & 19.2         \\
                  && FUNIT (ICCV'19)   & 34.0 & 94.6 & 31.8 && 22.9 & 21.6 & 22.2 && 0.147 & 19.2     \\
                  && LF-Font (AAAI'21) & 58.7 & 96.9 & 57.0 && 19.5 & 12.3 & 15.6  && \textbf{0.119} & \textbf{14.8}     \\
                  && \ours (proposed)  & \textbf{78.9} & \textbf{99.5} & \textbf{78.7} && \textbf{34.5} & 31.8 & \textbf{35.2} && 0.120 & 21.8  \\ \midrule
\parbox[t]{0mm}{\multirow{5}{*}{\rotatebox[origin=c]{90}{CN $\xrightarrow{}$ KR}}}
                  && EMD (CVPR'18)     & 4.6       & 15.4  & 0.8      && 0.8 & 0.1 & 0.1 && - & 150.1   \\
                  && AGIS-Net (TOG'19) & 13.3      & 32.1  & 3.1      && 1.8 & 0.6 & 0.6 && - & 146.5   \\
                  && FUNIT (ICCV'19)   & 11.3      & 66.4  & 6.6      && 12.0 & 17.3 & 9.1 && - & 176.0   \\
                  && LF-Font (AAAI'21) & 47.6      & 28.7  & 12.8     && 10.6 & 0.7 & 1.0 && - & 148.7   \\
                  && \ours (proposed)  & \textbf{66.3}  & \textbf{75.9}  & \textbf{50.0} && \textbf{74.6}   & \textbf{81.3}   & \textbf{89.2}   && - & \textbf{84.1}   \\ \bottomrule

\end{tabular}
\vspace{.5em}
\caption{\small {\bf Performance comparison on few-shot font generation scenario.} The performances of five few-shot font generation methods with four reference images are compared. We report accuracy measured by style-aware (Acc (S)) and content-aware (Acc (C)) classifiers and accuracy considering both the style and content labels (Acc (B)).
The summarized results of the user study are also reported. The User preference on considering style (User (S)), content (User (C)), both of them (User (B)) are shown.
LPIPS shows a perceptual dissimilarity between the ground truth and the generated glyphs. The harmonic mean (H) of style-aware and content-aware FID is also reported. Note that the FIDs are computed differently in two \ffg scenarios. All numbers are average of $50$ runs with different reference glyphs.}
\label{table:main_fewshot}
\vspace{-.5em}
\end{table*}

\begin{figure*}[t]
    \centering
    \includegraphics[width=\linewidth]{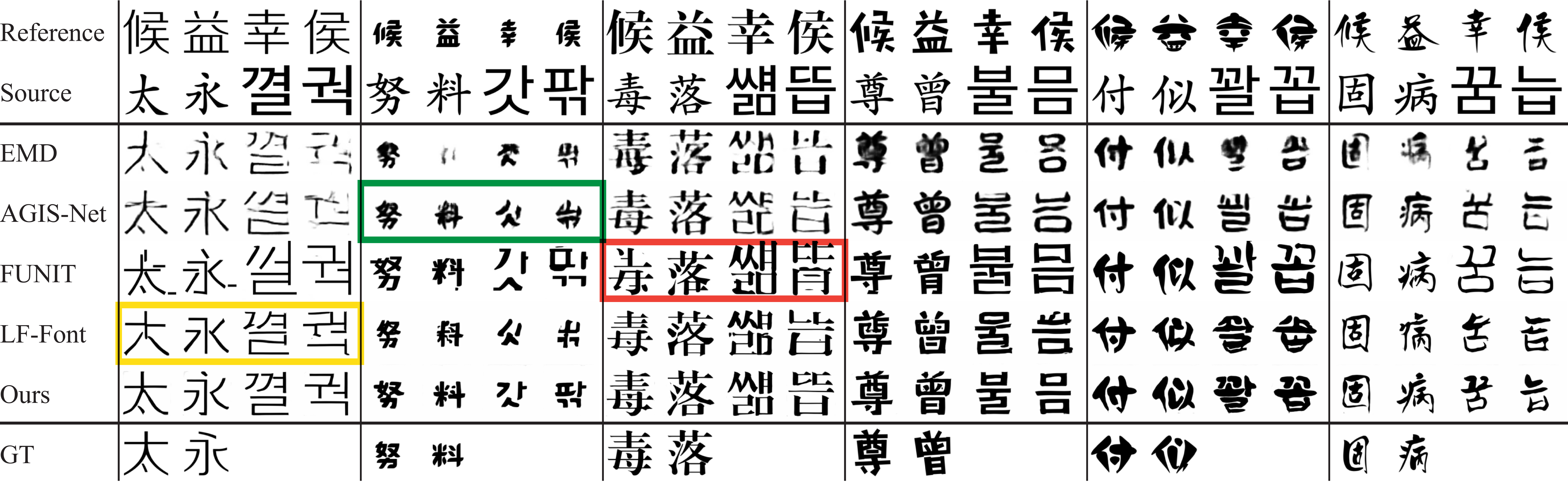}%
    \caption{\small \textbf{Generated Samples.} The generated images by five different models are shown. We also provide the reference and the source images used for the generation in the top two rows. The available ground truth images (GT) are shown in the bottom row. We highlight the samples that reveal the drawback of each model with colored boxes; green for AGIS-Net, red for FUNIT, and yellow for LF-Font.}%
    \label{fig:main_few}
    \vspace{-.5em}
\end{figure*}